# Learning Bayesian Nets that Perform Well

**Russell Greiner**
Siemens Corporate Research
755 College Road East
Princeton, NJ 08540-6632
greiner@scr.siemens.com

**Adam J. Grove**
NEC Research Institute
4 Independence Way
Princeton, NJ 08540
grove@research.nj.nec.com

**Dale Schuurmans***
Inst. for Research in Cognitive Science
University of Pennsylvania
Philadelphia, PA 19104-6228
daes@linc.cis.upenn.edu

## Abstract

A Bayesian net (BN) is more than a succinct way to encode a probabilistic distribution; it also corresponds to a function used to answer queries. A BN can therefore be evaluated by the accuracy of the answers it returns. Many algorithms for learning BNs, however, attempt to optimize another criterion (usually likelihood, possibly augmented with a regularizing term), which is independent of the distribution of queries that are posed. This paper takes the "performance criteria" seriously, and considers the challenge of computing the BN whose performance — read "accuracy over the distribution of queries" — is optimal. We show that many aspects of this learning task are more difficult than the corresponding subtasks in the standard model.

## 1 INTRODUCTION

Many tasks require answering questions; this model applies, for example, to both expert systems that identify the underlying fault from a given set of symptoms, and control systems that propose actions on the basis of sensor readings. When there is not enough information to answer a question with certainty, the answering system might instead return a probability value as its answer. Many systems that answer such probabilistic queries represent the world using a "Bayesian net" (BN), which succinctly encodes a distribution over a set of variables. Often the underlying distribution, which should be used to map questions to appropriate responses, is not known *a priori*. In such cases, if we have access to training examples, we can try to learn the model.

There are currently many algorithms for learning BNs [Hec95, Bun96]. Each such learning algorithm tries to determine which BN is optimal, usually based on some measure such as log-likelihood (possibly augmented with a "regularizing" term, leading to measures like MDL [LB94], and Bayesian Information Criterion (BIC) [Sch78]). However, these typical measures are independent of the queries that will be posed. To understand the significance of this, note that we may only care about certain queries (*e.g.*, the probability of certain specific diseases given a set of observed symptoms); and a BN with the best (say) log-likelihood given the sample may not be the one which produces the appropriate answers for the queries we care about. This paper therefore argues that BN-learning algorithms should consider the distribution ●f *queries*, as well as the underlying distribution of events, and should therefore seek the BN with the best performance *over the query distribution*, rather than the one that appears closest to the underlying event distribution.

To make this point more concrete, suppose we knew that all queries will be of the form $p(H \mid J, B)$ for some assignments to these variables (*e.g.*, *H*epatitis, given the possible symptoms *J*aundice=false and *B*lood test = true). Given a set of examples, our learner has to decide which BN (perhaps from some specified restricted set) is best. Now imagine we had two candidates BNs from this set: $B_1$, which performs optimally on the queries $p(H \mid J, B)$, but does horribly on other queries (*e.g.*, incorrectly claims that *J* and *B* are conditionally independent, has the completely wrong values for the conditional probability of *H* to the treatment *T* ("*T*ake aspirin"), and so on); versus $B_2$, which is slightly off on the $p(H \mid J, B)$ queries, but perfect on all other queries. Here, most measures would prefer $B_2$ over $B_1$, as they would penalize $B_1$ for its errors on the queries that will never occur! Of course, if we really do only care about $p(H \mid J, B)$, this $B_2$-over-$B_1$ preference is wrong.

This assumes we have the correct distributions, of both the real world events (*e.g.*, quantities like $p(H=1 \mid J = 0, B = 1) = 0.42$), and the queries that will be posed (*e.g.*, 48% of the queries will be of the form "What is $p(H = h \mid J = j, B = b)$?"; 10% will be "What is $p(H = h \mid S_1 = v_1, S_4 = v_4, S_7 = v_7)$?", etc.). Another more subtle problem with the maximal-likelihood–based measures arises when these distributions are not given explicitly, but must instead be estimated from examples. Here, we would, of course, like to use the given examples to obtain good estimates of the conditional probabilities $P(H|J, B)$. In the general maximal-likelihood framework, however, the examples would be used to fit *all* of the parameters within the entire BN, so we could conceivably "waste" some examples or computational effort learning the value of irrelevant parameters. In general, it seems better to focus the learner's resources on the relevant queries (but see Section 4).

Our general challenge is to acquire a BN whose performance is optimal, *with respect to the distribution of queries*, and the underlying distribution of events. Section 2 first lays out the framework by providing the relevant definitions. Section 3 then addresses several issues related to learning a BN whose accuracy (by this measure) is optimal: presenting the computational/sample complexities of first evaluating the quality of a given BN and then of finding the best BN of a given structure. It then provides methods for hill-climbing to a locally optimal BN. We will see that these tasks are computationally difficult for general classes of queries; Section 3 also presents a particular class of queries for which these tasks are easy. Section 4 then reflects on the general issue of how to best use knowledge of the query distribution to improve the efficiency of learning a good BN under our model. Here we show situations where this information may lead to ways of learning a BN (of a given structure) that are more sample-efficient than the standard approach. We first close this section by discussing how our results compare with others.

**Related Results:** The framework closest to ours is Friedman and Goldszmidt [FG96], as they also consider finding the BN that is best for some distribution of queries, and also explain why the BN with (say) maximal log-likelihood may not be the one with optimal performance on a specific task. In particular, they note that evaluating a Bayesian net $B$, given a set of training data $D = \{c^i, a_1^i, \ldots, a_n^i\}_{i=1}^N$, under the log-likelihood measure, amounts to using the formula

$$
\begin{aligned}
LL(B|D) &= \sum_{i=1}^N \log B(c^i | a_1^i, \ldots, a_n^i) \\
&+ \sum_{i=1}^N \log B(a_1^i, \ldots, a_n^i)
\end{aligned}
$$

where $B(\chi)$ is the probability that $B$ assigns to the event $\chi$. If all of the queries, however, ask for the value of $c$ given values of $\langle a_1, \ldots a_n \rangle$, then only the first summation matters. This means that systems that use $LL(B|D)$ to rank BNs could do poorly if the contributions of second summation dominate those of the first. The [FG96] paper, however, considers only building BNs for *classification*, i.e., where *every* query is of the specific form $p(C = c | A_1 = a_1, \ldots, A_n = a_n)$ where $C$ is the only "consequent" variable, and $\{A_i\}$ is the set of *all* other variables; their formulation also implicitly assumes that all possible query-instances (of complete tuples) will occur, and all are equally likely. By contrast, we do not constrain the set of queries to be of this single form, nor do we insist that all queries occur equally often, nor that all variables be involved in each query. Note in particular that we allow the query's antecedents to *not* include the Markov blanket of the consequent; we will see that this restriction considerably simplifies the underlying computation.

Each of the queries we consider is of the form "$p(\mathbf{X} = \mathbf{x} | \mathbf{Y} = \mathbf{y}) = ?$", where $\mathbf{X}, \mathbf{Y}$ are subsets of the variables, and $\mathbf{x}, \mathbf{y}$ are respective (ground) assignments to these variables. As such, they resemble the standard class of "statistical queries", discussed by Kearns and others [Kea93] in the context of noise-tolerant learners.[1] In that model, however, the *learner* is posing such queries to gather information about the underlying distribution, and the learner's score depends its accuracy with respect to some other specific set of queries (here the same $p(C = c | A_1 = a_1, \ldots, A_n = a_n)$ expression mentioned above). In our model, by contrast, the learner is observing which such queries are posed by the "environment", as it will be evaluated based on its accuracy with respect to these queries.

Other researchers, including [FY96, Höf93], also compute the sample complexity for learning good BNs. They, however, deal with likelihood-based measures, which (as we shall see) have some fundamental differences from our query-answering based model; hence, our results are incomparable.

## 2 FRAMEWORK

As a quick synopsis: a Bayesian net is a directed acyclic graph $\langle \mathcal{V}, E \rangle$, whose nodes represent variables, and whose arcs represent dependencies. Each node also includes a conditional-probability-table that specifies how the node's values depends (stochastically) on the values of its parents. (Readers unfamiliar with these ideas are referred to [Pea88].)

In general, we assume there is a stationary underlying distribution $P$ over the $N$ variables $\mathcal{V} = \{V_1, \ldots, V_N\}$. (I.e., $p(V_1 = v_1, \ldots, V_N = v_N) \geq 0$ and $\sum_{v_1, \ldots, v_N} p(V_1 = v_1, \ldots, V_N = v_N) = 1$. For example, perhaps $V_1$ is the "disease" random variable, whose value ranges over {healthy, cancer, flu, ...}; $V_2$ is "gender" $\in$ {male, female}, $V_3$ is "body_temperature" $\in$ [95..105], etc. We will refer to this as the "underlying distribution" or the "distribution over events".

A *statistical query* is a term of the form "$p(\mathbf{X} = \mathbf{x} | \mathbf{Y} = \mathbf{y}) = ?$", where $S, T \subset \mathcal{V}$ are (possibly empty) subsets of $\mathcal{V}$, and $\mathbf{x}$ (resp., $\mathbf{y}$) is a legal assignment to the elements of $\mathbf{X}$ (resp, $\mathbf{Y}$). We let $\mathcal{SQ}$ be the set of all possible legal statistical queries,[2] and assume there is a (stationary) distribution over $\mathcal{SQ}$, written $sq(\mathbf{X} = \mathbf{x} \,;\, \mathbf{Y} = \mathbf{y})$, where $sq(\mathbf{X} = \mathbf{x} \,;\, \mathbf{Y} = \mathbf{y})$ is the probability that the query "What is the value of $p(\mathbf{X} = \mathbf{x} \mid \mathbf{Y} = \mathbf{y})$?" will be asked. We of course allow $\mathbf{Y} = \{\}$; here we are requesting the prior value of $\mathbf{X}$, independent of any conditioning. We also write $sq(\mathbf{x}\,;\,\mathbf{y})$ to refer to the probability of the query "$p(\mathbf{X} = \mathbf{x} \mid \mathbf{Y} = \mathbf{y}) = ?$" where the variable sets $\mathbf{X}$ and $\mathbf{Y}$ can be inferred from the context.

---

[1] Of course, other groups have long been interested in this idea; *cf.*, the work in finding statistical answers from database queries.

While all of our results are based on such "ground" statistical queries, we could also define $sq(\mathbf{X}\,;\,\mathbf{Y})$ to refer to the probability that some query of the general form "$p(\mathbf{X} = \mathbf{x} \mid \mathbf{Y} = \mathbf{y}) = ?$" will be asked, for some assignments $\mathbf{x}, \mathbf{y}$; we could assume that all assignments to these unspecified variables are equally likely as queries. Finally, to simplify our notation, we will often use a single variable, say $q$, to represent the entire $[\mathbf{X} = \mathbf{x}, \mathbf{Y} = \mathbf{y}]$ situation, and so will write $sq(q)$ to refer to $sq(\mathbf{X} = \mathbf{x}\,;\,\mathbf{Y} = \mathbf{y})$.

As a simple example, the claim $sq(C\,;\,A_1, \ldots, A_n) = 1$ states that our BN will only be used to find classifications $C$ given the values of all of the variables $\{A_i\}$. (Notice this is *not* asserting that $p(C = c \mid A_1 = a_1, \ldots, A_n = a_n) = 1$ for any set of assignments $\{c, a_i\}$.) If all variables are binary, this corresponds to the claim that $sq(C = c\,;\,A_1 = a_1, \ldots, A_n = a_n) = 1/2^{n+1}$ for each assignment. Alternatively, we can use $sq(C\,;\,A_1, A_2, A_3) = 0.3$, $sq(C\,;\,A_1, A_2) = 0.2$, and $sq(D\,;\,C = 1,\,A_1 = 0,\,A_3) = 0.25$, and $sq(D\,;\,C = 1,\,A_1 = 1,\,A_3) = 0.25$, to state that 30% of the queries involve seeking the conditional probability of $C$ given the observed values of the 3 attributes $\{A_1, A_2, A_3\}$; 20% involve seeking the probability of $C$ given only the 2 attributes $\{A_1, A_2\}$; 25% seek the probability of the (different "consequent") $D$ given that $C = 1$, $A_1 = 0$ and some observed value of $A_3$, and the remaining 25% seek the probability of $D$ given that $C = 1$, $A_1 = 1$ and some observed value of $A_3$.

If each of the $N$ variables has domain $\{0, 1\}$, then the $\mathcal{SQ}$ distribution has $O(5^N)$ parameters, because each variable $v_i \in \mathcal{V}$ can play one of the following 5 roles in a query:

$$\left\{ \begin{array}{ccc} \left[\begin{array}{c} v \in \mathbf{X} \\ v = 1 \end{array}\right], & \left[\begin{array}{c} v \in \mathbf{X} \\ v = 0 \end{array}\right], & \left[\begin{array}{c} v \in \mathbf{Y} \\ v = 1 \end{array}\right], \\ \left[\begin{array}{c} v \in \mathbf{Y} \\ v = 0 \end{array}\right], & \left[\begin{array}{c} v \notin S \\ v \notin T \end{array}\right] & \end{array} \right\}$$

(We avoid degeneracies by assuming $\mathbf{Y} \cap \mathbf{X} = \{\}$.)

Notice that we assume that $sq(\,\cdot\,;\,\cdot\,)$ can be, in general, completely unrelated to $p(\,\cdot \mid \cdot\,)$, because the probability of being asked about $sq(\mathbf{X} = \mathbf{x}\,;\,\mathbf{Y} = \mathbf{y})$ need not be correlated (or at least, not in any simple way) with the value of the conditional probability $p(\mathbf{X} = \mathbf{x} \mid \mathbf{Y} = \mathbf{y})$. The fact that the underlying $p(\,\cdot\,)$ is stationary simply means that the query $sq(\,\cdot\,;\,\cdot\,)$ has a determinate answer given by the true conditional probability $p(\mathbf{X} = \mathbf{x} \mid \mathbf{Y} = \mathbf{y}) \in [0, 1]$. In general, we call each tuple $\langle \mathbf{X} = \mathbf{x},\,\mathbf{Y} = \mathbf{y};\, p(\mathbf{X} = \mathbf{x} \mid \mathbf{Y} = \mathbf{y}) \rangle$ a "labeled statistical query".

---

[2]A query $sq(\mathbf{X} = \mathbf{x}\,;\,\mathbf{Y} = \mathbf{y})$ is "legal" if $p(\mathbf{Y} = \mathbf{y}) > 0$. Note also that we use $CAPITAL$ letters to represent single variables, *lowercase* letters for that values that the variables might assume, and the **boldface** font when dealing with *sets* of variables or values.

Now fix a network (over $\mathcal{V}$) $B$, and let $B(\mathbf{x} \mid \mathbf{y}) = B(\mathbf{X} = \mathbf{x} \mid \mathbf{Y} = \mathbf{y})$ be the real-value (probability) that $B$ returns for this assignment. Given distribution $sq(\,\cdot\,;\,\cdot\,)$ over $\mathcal{SQ}$, the "score" of $B$ is

$$\mathrm{err}_{sq,p}(B) = \sum_{\mathbf{x}, \mathbf{y}} sq(\mathbf{x}\,;\,\mathbf{y})\,[B(\mathbf{x} \mid \mathbf{y}) - p(\mathbf{x} \mid \mathbf{y})]^2 \quad (1)$$

where the sum is over all assignments $\mathbf{x}, \mathbf{y}$ to all subsets $\mathbf{X}, \mathbf{Y}$ of variables. (We will often write this simply $\mathrm{err}(B)$ when the distributions $sq$ and $p$ are clear from the context.) Note this depends on both the underlying distribution $p(\,\cdot\,)$ over $\mathcal{V}$, and the $sq(\,\cdot\,)$ distribution over queries $\mathcal{SQ}$.

Given a set of labeled statistical queries $Q = \{\langle \mathbf{x}_i;\,\mathbf{y}_i;\,p_i \rangle\}_i$ we let

$$\overline{\mathrm{err}}^Q(B) = \frac{1}{|Q|} \sum_{\langle \mathbf{x};\,\mathbf{y};\,p \rangle \in Q} [B(\mathbf{x} \mid \mathbf{y}) - p]^2$$

be the "empirical score" of the Bayesian net.

For comparison, we will later use $\mathrm{KL}(B) = \sum_d p(d) \log \frac{p(d)}{B(d)}$ to refer to the Kullback-Liebler divergence between the correct distribution $p(\,\cdot\,)$ and the distribution represented by the Bayesian net $B(\cdot)$. Given a set $D$ of event tuples, we can approximate this score using $\overline{\mathrm{KL}}^D(B) = \frac{1}{|D|} \sum_{d \in D} \log \frac{1/|D|}{B(d)}$. Note (1) that small $KL$ divergence corresponds to the large (log)likelihood, and (2) that neither $\mathrm{KL}(B)$ nor $\overline{\mathrm{KL}}^D(B)$ depend on $sq(\,\cdot\,)$.

Finally, let $\mathcal{SQ}_B \subset \mathcal{SQ}$ be the class of queries whose "consequent" is single literal $\mathbf{X} = \{V\}$, and whose "antecedents" $\mathbf{Y}$ are a superset of $V$'s Markov blanket, with respect to the BN $B$; we will call these "Markov-blanket queries".

## 3   LEARNING ACCURATE BAYESIAN NETS

Our overall goal is to learn the Bayesian Net with the optimal performance, given examples of both the underlying distribution, and of the queries that will be posed (i.e., instances of $\{V_1 = v_1, \ldots, V_N = v_N\}$ tuples and instances of $\mathcal{SQ}$, possibly labeled).

**Observation 1** *Any Bayesian net $B_*$ that encodes the underlying distribution $p(\,\cdot\,)$, will in fact produce the optimal performance; i.e., $err(B_*)$ will be optimal. (However, the converse is not true: there could be nets* *whose performance is perfect on the queries that interest us, i.e., $err(B_*) = 0$, but which are otherwise very different from the underlying distribution.)* ∎

From this observation we see that, *if* we have a learning algorithm that produces better and better approximations to $p(\,\cdot\,)$ as it sees more training examples, then in the limit the $sq(\,\cdot\,)$ distribution becomes irrelevant.

Given a small set of examples, however, the $sq(\,\cdot\,)$ distribution can play an important role in determining which BN is optimal. This section considers both the computational and sample complexity of this underlying task. Subsection 3.1 first considers the simple task of evaluating a given network, as this information is often essential to learning a good BN. Subsection 3.2 then analyses the task of filling in the optimal CP-tables for a given graphical structure, and Subsection 3.3 discusses a hill-climbing algorithm for filling these tables, to produce a BN whose accuracy is locally optimal.

## 3.1   COMPUTING err($B$)

It is easy to compute the estimate $\overline{\mathrm{KL}}^D(B)$ of KL($B$), based on examples of *complete* tuples $D$ drawn from the $p(\,\cdot\,)$ distribution. In contrast, it is hard to compute the estimate $\overline{\mathrm{err}}^Q(B)$ of err($B$) from general statistical queries — in fact, it is not even easy to approximate this estimate.

**Theorem 2 ([Rot96, DL93])** *It is #P-hard[3] to compute $\overline{\mathrm{err}}^Q(B)$ over a set of general queries $Q \subset SQ$. It is NP-hard to even estimate this quantity to within an additive factor of 0.5.* ∎

The reason is that evaluating the score for an arbitrary Bayesian network requires evaluating the posterior probabilities of events in $Q$, which is known to be difficult in general. In fact, this is hard even if we know the distribution $p(\,\cdot\,)$ and consider only a single (known) form for the query.

Note, however, that this computation is much easier in the $SQ_B$ case, because there is an trivial way to evaluate a Bayesian net on any Markov-blanket query [Pea88]; and hence to compute the score.

There is an obvious parallel between estimating $\overline{\mathrm{err}}^{Q'}(B)$ when dealing with $SQ_B$ queries $Q'$, and estimating $\overline{\mathrm{KL}}^{D'}(B)$ from complete tuples $D'$ [Hec95]: both tasks are quite straightforward, basically because their respective Bayesian net computations are simple. Similarly, it can be challenging to compute $\overline{\mathrm{err}}^Q(B)$ in the general $SQ$ case, or to estimate $\overline{\mathrm{KL}}^D(B)$ from *incomplete* tuples $D$ [CH92, RBKK95]; as here the Bayesian net computations are inherently intractable. We will see these parallels again below.

---
[3]Roughly speaking, #P is the class of problems corresponding to *counting* the number of satisfiable assignments to a satisfiability problem, and thus #P-hard problems are at least as difficult as problems in NP.

Another challenge is computing the *sample complexity* of gathering the information required to compute the score for a network. It is easy to collect a sufficient number of examples if we are considering learning from *labeled* statistical queries. Here, a simple application of Hoeffding's Inequality [Hoe63] shows[4]

**Theorem 3** *Let*

$$\overline{\mathrm{err}}^Q(B) = \frac{1}{M_{LSQ}} \sum_{\langle q,p \rangle \in S_{LSQ}} (B(q) - p)^2$$

*be the empirical score of the Bayesian net $B$, based on a set $S_{LSQ}$ of*

$$M_{LSQ} = M_{LSQ}(\epsilon, \delta) = \frac{1}{2\epsilon^2} \ln \frac{2}{\delta}$$

*labeled statistical queries, drawn randomly from the $sq(\,\cdot\,)$ distribution and labeled by $p(\,\cdot\,)$. Then, with probability at least $1-\delta$, $\overline{\mathrm{err}}^{S_{LSQ}}(B)$ will be within $\epsilon$ of $err(B)$; i.e., $P[\,|\overline{\mathrm{err}}^{S_{LSQ}}(B) - err(B)| < \epsilon\,] \geq 1-\delta$, where this distribution is over all sets of $M_{LSQ}(\epsilon, \delta)$ randomly drawn statistical queries.* ∎

A more challenging question is: What if we only get *unlabeled* queries, together with examples of the underlying distribution? Fortunately, we again need only a polynomial number of (unlabeled) query examples. Unfortunately, we need more information before we can bound on the number of event examples required. To see this, imagine $sq(\,\cdot\,)$ puts all of the weight on a single query, i.e., $sq(X = 1; Y = 1) = 1$. Hence, a BN's accuracy depends completely on its performance on this query, which in turn depends critically on the true conditional probability $p(X = 1\,|\,Y = 1)$. The only event examples relevant to estimating this quantity are those with $Y = 1$; of course, these examples only occur with probability $p(Y = 1)$. Unfortunately, this probability can be arbitrarily small. Further, even if $p(Y = 1) \approx 0$, the true value of $p(X = 1\,|\,Y = 1)$ can still be large (e.g., if $X$ is equal to $Y$, then $p(X = 1\,|\,Y = 1) = 1$, even if $p(Y = 1) = 1/2^n$). Hence, we cannot simply ignore such queries (as $sq(\mathbf{X} = \mathbf{x}; \mathbf{Y} = \mathbf{y})$ can be high), nor can we assume the resulting value will be near 0 (as $p(\mathbf{X} = \mathbf{x}\,|\,\mathbf{Y} = \mathbf{y})$ can be high).

We can still estimate the score of a BN, in the following on-line fashion:

**Theorem 4** *First, let $S_{SQ} = \{sq(\mathbf{x}_i\,;\,\mathbf{y}_i)\}_i$ be a set of*

$$M_{SQ}(\epsilon, \delta) = \frac{2}{\epsilon^2} \ln \frac{4}{\delta}$$

---
[4]Proofs for all new theorems, lemmas and corollaries appear in [GGS97].

*unlabeled statistical queries drawn randomly from the* $sq(\cdot;\cdot)$ *distribution. Next, let* $S_D$ *be the set of (complete) examples sequentially drawn from the underlying distribution* $p(\cdot)$, *until it includes at least*

$$M'_D(\epsilon,\delta) \;=\; \frac{8}{\epsilon^2}\ln\frac{2\,M_{SQ}}{\delta}$$

*instances that match each* $\mathbf{y}_i$ *value; notice* $S_D$ *may require many more than* $M'_D$ *examples. (The "legal query" requirement* $p(\mathbf{y}_i) > 0$ *insures that this collection process will terminate, with probability 1.) Finally, let* $\hat{p}^{(S_D)}(\mathbf{x}_i \,|\, \mathbf{y}_i)$, *be the empirically observed estimate of* $p(\mathbf{x}_i \,|\, \mathbf{y}_i)$, *based on this* $S_D$ *set. Then, with probability at least* $1-\delta$,

$$\overline{err}^{S_{SQ},S_D}(B) = \frac{1}{|S_{SQ}|}\sum_{\langle \mathbf{x},\mathbf{y}\rangle \in S_{SQ}}\Big[B(\mathbf{x}\,|\,\mathbf{y}) - \hat{p}^{(S_D)}(\mathbf{x}|\mathbf{y})\Big]^2$$

*will be within* $\epsilon$ *of* $err(B)$; *i.e.,* $P[\,|\overline{err}^{S_{SQ},S_D}(B) - err(B)| < \epsilon\,] \;\geq\; 1-\delta$. ∎

We can, moreover, get an *a priori* bound on the total number of event examples *if we can bound the probability of the query's conditioning events*. That is,

**Corollary 5** *If we know that all queries encountered,* $sq(\mathbf{x};\mathbf{y})$, *satisfy* $p(\mathbf{y}) \geq \lambda$ *for some* $\lambda > 0$, *then we need only gather*

$$M_D(\epsilon,\delta,\lambda) \;=\; \max\left\{\;\frac{2}{\lambda}\left[M'_D + \ln\frac{4M_{SQ}}{\delta}\right],\; \frac{8}{\epsilon^2}\ln\frac{4\,M_{SQ}}{\delta}\;\right\}$$

*complete event examples, along with*

$$M_{SQ}(\epsilon,\delta) \;=\; \frac{2}{\epsilon^2}\ln\frac{4}{\delta}$$

*example queries, to obtain an* $\epsilon$-*close estimate, with probability at least* $1-\delta$. ∎

Of course, as $\lambda$ can be arbitrarily small (e.g., $o(1/2^n)$ or worse), this $M_D$ bound can be arbitrarily large, in terms of the size of the Bayesian net. Note also that the Friedman and Yakhini [FY96] bound similarly depends on "skewness" of the distribution, which they define as the smallest non-zero probability of an event, over all atomic events.[5]

Two final comments: (1) Recall that these bounds describe only *how many examples are required*; not how much work is required, given this information. Unfortunately, using these examples to compute the score of a BN requires solving a $\#P$-hard problem; see Theorem 2. (2) The sample complexity results hold for estimating the accuracy of any system for representing arbitrary distributions; not just BNs.

---

[5]Höffgen [Höf93] was able to avoid this dependency, in certain "log-loss" contexts, by "tilting" the empirical distribution to avoid 0-probability atomic events. That trick does not apply to our query-based error measure.

## 3.2 COMPUTING OPTIMAL CP-tables FOR A GIVEN NETWORK STRUCTURE

The structure of a Bayesian net, in essence, specifies which variables are directly related to which others. As people often know this "causal" information (at least approximately), many BN-learners actually begin with a given structure, and are expected to use training examples to "complete" the BN, by filling in the "strength" of these connections — *i.e.*, to learn the CP-table entries. To further motivate this task of "fitting" good CP-tables to a given BN structure, note that it is often the key sub-routine of the more general BN-learning systems, which must also search through the space of structures. This subsection addresses both the computational, and sample, complexity of finding this best (or near best) CP-table. Subsection 3.3 next suggests a more practical, heuristic approach.

Stated more precisely, the *structure* of a specific Bayesian net is a directed acyclic graph $\langle \mathcal{V},E\rangle$ with nodes $\mathcal{V}$ and edges $E \subset \mathcal{V}\times\mathcal{V}$. There are, of course, (uncountably) many BNs with this structure, corresponding to all ways of filling in the CP-tables. Let $\mathcal{BN}(\mathcal{V},E)$ denote all such BNs.

We now address the task of finding a BN $B \in \mathcal{BN}(\mathcal{V},E)$ whose score is, with high probability, (near) minimal among this class; *i.e.*, find $B$ such that

$$err(B) \;\leq\; \epsilon + \min_{B'\in\mathcal{BN}(\mathcal{V},E)}err(B')$$

with probability at least $1-\delta$, for small $\epsilon,\delta > 0$. As in Subsection 3.1, our learner has access to either *labeled* statistical queries drawn from the query distribution $sq(\cdot)$ over $S\mathcal{Q}$; or unlabeled queries from $sq(\cdot)$, together with event examples drawn from $p(\cdot)$.

Unfortunately this task — like most other other interesting questions in the area — appears computationally difficult in the worst case. In fact, we prove below the stronger result that finding the (truly) optimal Bayesian net is not just NP-hard, but is actually non-approximatable:

**Theorem 6** *Assuming* $P \neq NP$, *no polynomial-time algorithm (using only labeled queries) can compute the CP-tables for a given Bayesian net structure whose error score is within a sufficiently small additive constant of optimal. That is, given any structure* $\langle \mathcal{V},E\rangle$ *and a set of labeled statistical queries* $Q$, *let* $B_{\langle \mathcal{V},E\rangle,Q} \in \mathcal{BN}(\mathcal{V},E)$ *have the minimal error over* $Q$; *i.e.,* $\forall B' \in \mathcal{BN}(\mathcal{V},E)$, $\overline{err}^Q(B_{\langle \mathcal{V},E\rangle,Q}) \;\leq\; \overline{err}^Q(B')$. *Then (assuming* $P \neq NP$) *there is some* $\gamma > 0$ *such that no polynomial-time algorithm can always find a solution within* $\gamma$ *of optimal, i.e., no polytime algorithm can always return a* $B''_{\langle \mathcal{V},E\rangle,Q}$ *such that* $\overline{err}^Q(B''_{\langle \mathcal{V},E\rangle,Q}) - \overline{err}^Q(B_{\langle \mathcal{V},E\rangle,Q}) \leq \gamma$. ∎

In contrast, notice that the analogous task is trivial in the log-likelihood framework: Given complete training examples (and some benign assumptions), the CP-table that produces the optimal maximal-likelihood BN corresponds simply to the observed frequency estimates [Hec95].

However, the news is not all bad in our case. Although the problem may be computationally hard, the *sample* complexity can be polynomial. That is (under certain conditions; see below), if we draw a polynomial number of labeled queries, and (somehow!) find the BN $B$ that gives minimal error for those queries, then with high probability $B$ will be within $\epsilon$ of optimal over the full distribution $sq(\cdot)$.

We conjecture that the sample complexity result is true in general. However, our results below uses the following annoying, but extremely benign, technical restriction. Let $\mathcal{T} = \{ \mathbf{y} \mid sq(\mathbf{x}; \mathbf{y}) > 0 \}$ be the set of all conditioning events that might appear in queries (often $\mathcal{T}$ will simply be the set of all events). For any $c > 1$, define

$$BN_{\mathcal{T} \succeq 1/2^{c^N}}(\mathcal{V}, E) \;=\;$$
$$\{ B \in BN(\mathcal{V}, E) \mid \forall \mathbf{y} \in \mathcal{T}, B(\mathbf{y}) > 1/2^{c^N} \}$$

to be the subset of BNs that assign, to each conditioning event, a probability that is bounded below by the doubly-exponentially small number $1/2^{c^N}$. (Recall that $N = |\mathcal{V}|$, the number of variables.) We now restrict our attention to these Bayesian nets.[6]

**Theorem 7** *Consider any Bayesian net structure* $\langle \mathcal{V}, E \rangle$, *requiring the specification of* $K$ *CP-table entries* $CPT = \{[q_i | r_i]\}_{i=1..K}$. *Let* $B^* \in BN_{\mathcal{T} \succeq 1/2^{c^N}}(\mathcal{V}, E)$ *be the BN that has minimum empirical score with respect to a sample of*

$$M'_{LSQ}(\epsilon, \delta) \;=\;$$
$$\frac{1}{4\epsilon^2} \left( \log \frac{2}{\delta} + K \log \frac{2K}{\epsilon} + NK \log(2 + c - \log \epsilon) \right)$$

*labeled statistical queries from* $sq(\cdot)$. *Then, with probability at least* $1 - \delta$, $B^*$ *will be no more than* $\epsilon$ *worse than* $B^{opt}$, *where* $B^{opt}$ *is the BN with optimal score among* $BN_{\mathcal{T} \succeq 1/2^{c^N}}(\mathcal{V}, E)$ *with respect to the full distribution* $sq(\cdot)$. ∎

This theorem is nontrivial to prove, and in particular is *not* an immediate corollary to Theorem 3. That earlier result shows how to estimate the score for a single *fixed* BN, allowing $\delta$ probability of error. But since $B^*$ is chosen after the fact (*i.e.*, to be optimal on the training set) we cannot have the same confidence that we have estimated its score correctly. Instead, we must use sufficiently many examples so that the *simultaneously estimated scores* for *all* (uncountably many) $B' \in BN_{\mathcal{T} \succeq 1/2^{c^N}}(\mathcal{V}, E)$ are all within $\epsilon$ of the true values, with collective probability at most $\delta$ that there is *any* error. ●nly then can we be confident about $B^*$'s accuracy. (See proof in [GGS97].)

As in Section 3.1, we can also consider the slightly more complex task of learning the CP-table entries from *unlabeled* statistical queries $sq(\mathbf{X} = \mathbf{x}; \mathbf{Y} = \mathbf{y})$, augmented with examples of the underlying distribution $p(\cdot)$. However, as above, this is a straightforward extension of the "learning from labeled statistical query" case: one first draws a slightly larger sample of *un*labeled statistical queries, and then uses a sufficient sample of domain tuples to accurately estimate the labels for each of these queries (hence simulating the effect of drawing fewer — but still sufficiently many — *labeled* statistical queries). Here we encounter the same caveats that each of the unlabeled statistical queries $sq(\mathbf{X} = \mathbf{x}; \mathbf{Y} = \mathbf{y})$ must involve conditioning events $\mathbf{Y} = \mathbf{y}$ that occur with some nontrivial probability $p(\mathbf{Y} = \mathbf{y}) > 0$ (for otherwise one could not put an nontrivial upper bound on the number of tuples needed to learn a good setting of the CP-table entries). A detailed statement and proof of this result is a straightforward extension of Theorem 7, so we omit the details here. (See [GGS97].)

The point is that, from a sample complexity perspective, it *is* feasible to learn near optimal settings for the CP-table entries in a fixed Bayesian network structure under our model. The only difficult part is that actually *computing* these optimal entries from (a polynomial number of) training samples is hard in general; *cf.*, Theorem 6. In fact, we will see, in Section 4, that it is *not* correct to simply fill each CP-table entry with the frequency estimates.

## 3.3  HILL CLIMBING

It should not be surprising that finding the optimal CP-tables was computationally hard, as this problem has a lot in common with the challenge of learning the KL($\cdot$)-best network, given *partially specified tuples*; a task for which people often use iterative steepest-ascent climbing methods [RBKK95]. We now briefly consider the analogous approach in our setting.

Given a single labeled statistical query "$\langle \mathbf{x}; \mathbf{y}; p(\mathbf{x} \mid \mathbf{y}) \rangle$", consider how to change the value of the CP-table entry $[q|r]$, whose current value is $e_{q|r}$. We use the following lemma:

**Lemma 8** *Let* $B$ *be a Bayesian net whose CP-table includes the value* $e_{Q=q|R=r} = e_{q|r} \in [0, 1]$ *as the value for the conditional probability of* $Q = q$ *given* $\mathbf{R} = \mathbf{r}$. *Let* $sq(\mathbf{X}; \mathbf{Y})$ *be a statistical query, to which* $B$ assigns the probability $B(\mathbf{X} \mid \mathbf{Y})$. Then the derivative of $B(\mathbf{X} \mid \mathbf{Y})$, wrt the value $e_{q|\mathbf{r}}$, is

$$\frac{d B(\mathbf{X} \mid \mathbf{Y})}{d e_{q|\mathbf{r}}}$$
$$= \frac{1}{e_{q|\mathbf{r}}} B(\mathbf{X} \mid \mathbf{Y}) [B(q, \mathbf{r} \mid \mathbf{X}, \mathbf{Y}) - B(q, \mathbf{r} \mid \mathbf{Y})] \quad (2)$$

---

[6]Conceivably — although we conjecture otherwise — there could be some sets of queries and some graphs $\langle \mathcal{V}, E \rangle$, such that the best performance is obtained with extremely small CP-table entries; *e.g.*, of order $o(1/2^{2^{2^N}})$. (But note that numbers this small can require doubly-exponential precision just to write down, so such BNs would perhaps be impractical anyway!) Our result assumes that, even if such BNs do allow improved performance, we are not interested in them.

As $B$ produced the score $B(\mathbf{X} \mid \mathbf{Y})$ here, the error for this single query is $\overline{\mathrm{err}}^{(\mathbf{X}, \mathbf{Y})}(B) = (B(\mathbf{X} \mid \mathbf{Y}) - p)^2$. To compute the gradient of this error value, as a function of this single CP-table entry (using Equation 2),

$$\frac{d \overline{\mathrm{err}}^{(\mathbf{X}, \mathbf{Y})}(B)}{d e_{q|\mathbf{r}}} = 2(B(\mathbf{X} \mid \mathbf{Y}) - p) \frac{d B(\mathbf{X} \mid \mathbf{Y})}{d e_{q|\mathbf{r}}}$$

thus, letting $C = 2(B(\mathbf{X} \mid \mathbf{Y}) - p)$, we get

$$C \frac{d B(\mathbf{X} \mid \mathbf{Y})}{d e_{q|\mathbf{r}}}$$
$$= \frac{C}{e_{q|\mathbf{r}}} [B(q, \mathbf{r}, \mathbf{X} \mid \mathbf{Y}) - B(\mathbf{X} \mid \mathbf{Y}) B(q, \mathbf{r} \mid \mathbf{Y})]$$
$$= \frac{C}{e_{q|\mathbf{r}}} B(\mathbf{X} \mid \mathbf{Y}) [B(q, \mathbf{r} \mid \mathbf{X}, \mathbf{Y}) - B(q, \mathbf{r} \mid \mathbf{Y})] \quad (3)$$

We can then use this derivative to update the $e_{q|\mathbf{r}}$ value, by hill-climbing in the direction of the gradient (i.e., gradient ascent.). Of course, Equation 3 provides only that component of the gradient derived from a single query; the overall gradient for $e_{q|\mathbf{r}}$ will involve summing these values of all queries (or perhaps all queries in some sample). Furthermore, we must constrain the gradient ascent to only move such that $\sum_{q}' e_{q'|\mathbf{r}}$ remains as 1 (i.e., the sum of probabilities of all possible values for $Q$, given a particular valuation for $Q$'s parents, must sum to 1). However, the techniques involved are straightforward and well-known, so we omit further analysis here.

Notice immediately from Equation 3 that we will not update $e_{q|\mathbf{r}}$ (at least, not because of the query $sq(\mathbf{X} ; \mathbf{Y})$) if the difference $B(\mathbf{X} \mid \mathbf{Y}) - p$ is 0 (i.e., if $B(\mathbf{X} \mid \mathbf{Y})$ is correct) or if $B(q, \mathbf{r} \mid \mathbf{X}, \mathbf{Y}) - B(q, \mathbf{r} \mid \mathbf{Y})$ is 0 (i.e., if $\mathbf{Y}$ "d-separates" $\mathbf{X}$ and $q, \mathbf{r}$); both of which makes intuitive sense.

Unfortunately, we see that evaluating the gradient requires computing conditional probabilities in a BN. This is analogous to to the known result in the traditional model [RBKK95]. It thus follows that it can be $\#P$-hard to evaluate this gradient in general (see Theorem 2). However, in special cases — i.e., BNs for which inference is tractable — efficient computation is possible.

One demonstration of this is the class of "Markov-blanket queries" $SQ_B$ (recall the definition at the end of Section 2). Carrying out the gradient computation is easy in this case: Here when updating the $[q|\mathbf{r}]$ entry, we can ignore queries $sq(\mathbf{X} ; \mathbf{Y})$ if $[q|\mathbf{r}]$ is outside of its Markov blanket. We therefore need only consider the queries $sq(\mathbf{X} ; \mathbf{Y})$ where $\{Q\} \cup \mathbf{R} \subset \mathbf{X} \cup \mathbf{Y}$ and moreover, when $Q = q$ is consistent with $\mathbf{X}$'s assignment; for these queries, the gradient is

$$\frac{d \overline{\mathrm{err}}^{(\mathbf{X}, \mathbf{Y})}(B)}{d e_{q|\mathbf{r}}}$$
$$= \frac{2(B(\mathbf{X} \mid \mathbf{Y}) - p)}{e_{q|\mathbf{r}}} B(\mathbf{X} \mid \mathbf{Y}) (1 - B(\mathbf{X} \mid \mathbf{Y})) (4)$$

which follows from Equation 3 using $B(q, \mathbf{r} \mid \mathbf{X}, \mathbf{Y}) = 1$ as $q, \mathbf{r}$ is consistent with $\mathbf{X}$'s claim (recall we ignore $sq(\mathbf{X} ; \mathbf{Y})$ otherwise), and observing that $B(q, \mathbf{r} \mid \mathbf{Y})$ reduces to $B(\mathbf{X} \mid \mathbf{Y})$, as the part of $\{Q\} \cup \mathbf{R}$ already in $\mathbf{Y}$ is irrelevant. Notice Equation 4 is simple to compute, as it involves no non-trivial Bayesian net computation; see the simple algorithms in [Pea88].

## 4    HOW CAN THE QUERY DISTRIBUTION HELP?

Our intuition throughout this paper is that having access to the distribution of queries should allow us to learn better and more efficiently than if we only got to see domain tuples alone. Is this really true?

Note that the simplest and most standard approach to learning CP-table entries is simply filling in each CP-table entry with the observed frequency estimates [OFE] obtained from $p(\cdot)$. Note that this ignores any information about the query distribution. Unfortunately, OFE is not necessarily a good idea in our model, even if we have an arbitrary number of examples. This follows immediately from Theorem 6: If the standard OFE algorithm was always successful, then we would have a trivial polynomial time algorithm for computing a near-optimal CP-table for a fixed Bayesian net structure — which cannot be (unless $P = NP$). Yes, Observation 1 does claim that the optimal BN is a faithful model of the event distribution, meaning in particular that the value of each CP-table entry $[q|\mathbf{r}]$ can be filled with the true probability $p(q \mid \mathbf{r})$. However, this claim is not true in general in the current context, where we are seeking the best CP-table entries for a given network structure, as this network structure might not correspond to the true conditional independence structure of the underlying distribution $p(\cdot)$.

In the case where the BN structure does not correspond to the true conditional independence structure of the underlying $p(\cdot)$, ignoring the query distribution and using straight OFE can lead to arbitrarily bad results:

**Example 4.1** Suppose the BN structure is simply $A \rightarrow X \rightarrow C$, and the only labeled queries are $\langle C; A; 1.0 \rangle$ and $\langle C; \bar{A}; 0.0 \rangle$. (I.e., $A \equiv C$ with probability 1, and we are only interested in querying $C$ given $A$ or $\bar{A}$.) Suppose further that the intervening $X$ is completely independent of $A$ and $C$ — i.e., $p(X \mid A) = p(X \mid \neg A) = p(C \mid X) = p(C \mid \neg X) = 0.5$. (Note that this BN structure is seriously wrong.)

*In this situation, the BN that most faithfully follows the event distribution, $B_p$, would have CP-table entries $e_{X|A} = e_{X|\bar{A}} = e_{C|X} = e_{C|\bar{X}} = 0.5$, with a performance score of $err(B_p) = 0.25$. (Recall that $e_{X|A}$ is the CP-table entry that specifies the probability of $X$, given that $A$ holds; etc.) Now consider $B_{sq}$, whose entries are $e_{X|A} = e_{C|X} = 1.0$ and $e_{X|\bar{A}} = e_{C|\bar{X}} = 0.0$ — i.e., make $X \equiv A$ and $C \equiv X$. While $B_{sq}$ clearly has the $X$-dependencies completely wrong, its score is perfect, i.e., $err(B_{sq}) = 0.0$.[7]* ∎

Thus, filling CP-table entries with observed frequency estimates — or using any other technique that converges to $B_p$ — leads to a bad solution in this case, no matter how many training examples are used. On the other hand, consider a learning procedure that (knowing the query distribution!) ignores the $X$ variable completely and directly estimates the conditional probabilities $p(A \mid C)$ and $p(A \mid \neg C)$ before filling in the CP-table entries (i.e., which is isomorphic to $B_{sq}$). This would eventually learn a perfect classifier. Of course, such a learning procedure might have to be based on the (impractical) learning techniques developed in Theorem 7, or perhaps (more realistically) use the heuristic hill-climbing strategies presented in Section 3.3.

What about the case when the proposed network structure *is* correct? Here we know that the standard OFE approach eventually *does* converge to an optimal CP-table setting for *any* query distribution (Observation 1). So, unlike the case of an incorrect structure, there is no reason in the large-sample-size limit to consider the query distribution. But what about the more realistic situation, where the sample is finite? The question then is:

Given that the known structure is correct, can we exploit knowing the true query distribution?

There is one simple sense in which the answer can certainly be yes. It is clearly safe to to restrict our attention to those nodes of the BN that are not d-separated from every query variable by the conditioning variables that appear in the queries. That is, if a BN $B$ contains the edge $U \to V$, and the query distribution $sq(\mathbf{X} = \mathbf{x}; \mathbf{Y} = \mathbf{y})$ is such that, for every query "$p(\mathbf{X} = \mathbf{x} \mid \mathbf{Y} = \mathbf{y}) = ?$", both $U$ and $V$ are d-separated from $\mathbf{X}$ by $\mathbf{Y}$, then we know that the CP-table entry $e_{v|u}$ cannot affect $B$'s answer to the query, $B(\mathbf{X} \mid \mathbf{Y})$. Thus, it seems clear that we do not need to bother estimating $e_{v|u}$ here. Now suppose we have a learning algorithm that uses a computed sample size bound (which grows with the number of parameters to be estimated) in order to provide certain performance guarantees. Here, our knowledge of the query distribution will reduce the effective size of the BN, which will allow us to stop learning after fewer samples. Thus, using the query distribution can give an advantage here, although only in a rather weak sense: the basic learning technique might still amount to filling in the CP-table entries with frequency estimates obtained from the underlying distribution $p(\cdot)$ — the only win is that we will know that it is safe to stop earlier because a small fragment of the network is relevant.

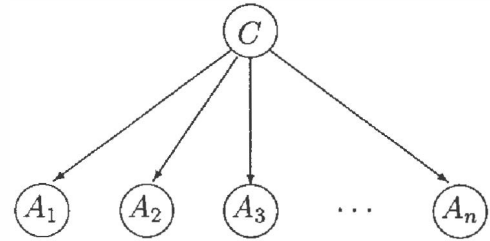

Figure 1: Bayesian network structure for Example 4.2 ("Naïve Bayes").

Can one do better than simply filling in CP-table entries with frequency estimates, given that the BN structure is correct? As we now show, this question does not seem to have a simple answer.

Motivated by Example 4.1, one might ignore the BN-structure in general, and just directly estimate the conditional probabilities for the queries of interest. Note that this is guaranteed to converge to an optimal solution, *eventually*, even if the BN structure is incorrect. However, it can be needlessly inefficient in some cases, especially if the postulated BN structure is known to be correct. This is because the BN structure can provide valuable knowledge about the distribution.

**Example 4.2** *Consider the standard "Naïve Bayes" model with $n + 1$ binary attributes $C, A_1, \ldots, A_n$ such that the $A_i$ are conditionally independent given $C$; see Figure 1. Suppose the single query of interest is "$p(C = 0 \mid A_1 = 0, A_2 = 0, \ldots, A_n = 0) = ?$". If we attempt to learn this probability directly, we must wait for the (possibly very rare) event that $A_1 = A_2 = \ldots = A_n = 0$; it is easy to construct situations where this will require an exponential (expected) number of examples. However, if we use the BN-structure, we can compute the required probability as soon as we have learned the $2n + 1$ probabilities $p(C = 0)$ and* $p(A_i = 0 | C = 0), p(A_i = 0 | C = 1)$ *for all* $i$. *If* $p(C = 0)$ *is near* $1/2$, *these probabilities will all be learned accurately after relatively few samples.* ∎

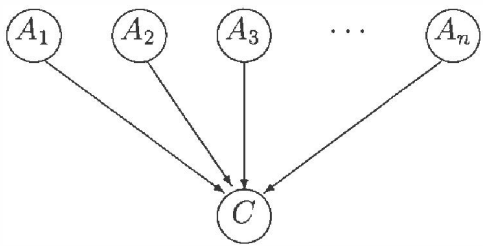

Figure 2: Bayesian network structure for Example 4.3 ("Reverse Naive Bayes").

This example might suggest that we should always try to learn the BN as accurately as possible, and ignore the query distribution (*e.g.*, just use OFE). However, there are other examples in which this approach would hurt us:

**Example 4.3** *Consider the "reverse" BN-structure from Example 4.2, where the arrows are now directed* $A_i \rightarrow C$ *instead of* $C \rightarrow A_i$ *(Figure 2), and assume we are only interested in queries of the form "*$p(C | \{\}) =$ ?*". Here the strategy that estimates* $p(C = c)$ *directly (hence, ignoring the given BN-structure) dominates the standard approach of estimating the CP-table entries. To see this, note that for any reasonable training sample size* $N \ll 2^n$, *the frequency estimates for most of the* $2^n$ *CP-table entries* $e_{c | a_1, \ldots, a_n}$ *will be undefined. Even using Laplace adjustments to compensate for this, the accuracy of the resulting BN estimate* $B(C = c | \{\})$ *will be poor unless* $p(C = c)$ *happens to be near 0.5. So, for example, if the true distribution* $p(\cdot)$ *is such that* $C \equiv A_1 \wedge parity(A_2, \ldots, A_n)$, *and* $p(A_i = 1) = 0.5$, *then* $p(C = 1)$ *will be equal to 0.25. But here the BN estimator (using Laplace adjustments) will give a value of* $B(C = 1 | \{\}) \approx 0.5$ *for any training sample size* $N \ll 2^n$, *whereas the direct strategy will quickly converge to an accurate estimate* $\hat{P}(C = 1) \approx 0.25$. *(Note that the standard maximum likelihood BN estimator will not even give a defined value for* $B(C = 1 | \{\})$ *in this case, since most of the possible* $a_1, \ldots, a_n$ *patterns will remain unobserved.)*

In general, the question

> What should we actually do if the BN structure is known (or assumed) to be correct, and we are training on a (possibly small) sample of complete instances?

remains open, and is an interesting direction for future work — asking, in essence, should we trust the given BN-structure, or the given query distribution? The previous two examples suggest that the answer is not a trivial one.

## 5    CONCLUSIONS

**Remaining Challenges:** There are of course several other obvious open questions.

First, the analyses above assume that we had the BN-structure, and simply had to fill in the values of the CP-tables. In general, of course, we may have to use the examples to learn that structure as well. The obvious approach is to hill-climb in the discrete, but combinatorial space of BN structures, perhaps using a system like PALO [Gre96], after augmenting it to climb from one structure $S_i$ to a "neighboring" structure $S_{i+1}$, if $S_{i+1}$, *filled with some CP-table entries*, appears better than $S_i$ with (near) optimal CP-table values, over a distribution of queries. Notice we can often save computation by observing that, for any query $q$, $B_1$ and $B_2$ will give the same error scores $B_1(q) = B_2(q)$ if the only differences between $B_1$ and $B_2$ are outside of $q$'s Markov blanket.

The second challenge is how best to accommodate *both* types of examples: queries (possibly labeled), and domain tuples. As discussed above, $sq(\cdot)$ examples are irrelevant *given complete knowledge of* $p(\cdot)$ (Observation 1). Similarly, *given complete knowledge of the query distribution*, we only need $p(\cdot)$ information to provide the labels for the queries.

Of course, these extreme conditions are seldom met; in general, we only have partial information of either distribution. Further, Example 4.1 illustrates that these two corpora of information may lead to different BNs. We therefore need some measured way of combining both types of information, to produce a BN that is both appropriate for the queries that have been seen, and for other queries that have not — even if this means degrading the performance on the observed queries. (That is, the learner should not "overfit" the learned BN to just the example queries it has seen; it should be able to "extend" the BN based on the event distribution.)

**Contributions:** As noted repeatedly in Machine Learning and elsewhere, the goal of a learning algorithm should be to produce a "performance element" that will work well on its eventual performance task [SMCB77, KR94]. This paper considers the task of learning an effective Bayesian net within this framework, and argues that the goal of a BN-learner should be to produce a BN whose error, over the distribution of queries, is minimal.

Our results show that many parts of this task are, unfortunately, often harder than the corresponding tasks when producing a BN that is optimal in more familiar contexts (*e.g.*, maximizing likelihood over the sampled event data) — see in particular our hardness results for evaluating a BN by our criterion (Theorem 2), for filling in a BN-structure's CP-table (Theorem 6), and for the steps used by the obvious hill-climbing algorithm trying to instantiate these tables. (Note, however, that our approach is robustly guaranteed to converge to a BN with optimal performance, while those alternative techniques are not.) Fortunately, we have found that the sample requirements are not problematic for our tasks (see Theorems 3, 4, 7 and Corollary 5), given various obvious combination of example types; we also identify a significant subclass of queries ($\mathcal{SQ_B}$) in which some of these tasks are computationally easy. We have also compared and contrasted our proposed approach to filling in the CP-table-entries with the standard "observed frequency estimate" method, and found that there are many subtle issues in deciding which of these "algorithms" works best, especially in the small-sample situation. These results are summarized in Table 1. We plan further analysis (both theoretical and empirical) towards determining when this more standard measure, now seen to be computationally simpler, is in fact an appropriate approximation to our performance-based criteria.

| Learning "Algorithm" | Structure is | Computational Efficiency | Correct convergence (in the limit) | Small sample |
|---|---|---|---|---|
| OFE[*] | Correct | "easy" | Yes [Obs 1] | $>^?$QD [Ex 4.2][**] |
|  | Incorrect |  | No [Ex 4.1] |  |
| QD[†] | Correct | $NP$-hard to approx. [Th 6] | Yes | $>^?$OFE [Ex 4.3][**] |
|  | Incorrect |  | Yes[‡] |  |

[*]  OFE = Observed Frequency Estimates (or any other algorithm that tries to match the event distribution.)

[†]  QD uses the CP-table that is "best" for given query distribution, using samples from the distribution to label queries.

[‡]  QD will produce the BN that has minimum error, for this structure.

[**]  Our examples illustrate cases in which one "algorithm" (OFE, QD) is more sample efficient than the other.

Table 1: Issues when Learning from Distributional Samples